\documentclass[journal,twoside,web]{ieeecolor}
\usepackage{generic}
\usepackage{cite}
\usepackage{amsmath,amssymb,amsfonts}
\usepackage{algorithmic}
\usepackage{graphicx}
\usepackage{textcomp}

\def\BibTeX{{\rm B\kern-.05em{\sc i\kern-.025em b}\kern-.08em
    T\kern-.1667em\lower.7ex\hbox{E}\kern-.125emX}}

\begin{document}
\title{Compensation Effect Amplification Control (CEAC): A movement-based approach for coordinated position and velocity control of the elbow of upper-limb prostheses}
\author{Julian Kulozik$^1$, Nathanaël Jarrassé$^1$
\thanks{$^1$Sorbonne Université, CNRS, INSERM, Institute for Intelligent Systems and Robotics (ISIR), Paris, France. }
\thanks{This research was partly funded through the Agence Nationale de la Recherche support, under grant~\texttt{\#}ANR-23-DMRO-0022.  Email: kulozik@isir.upmc.fr}
}
\maketitle

\begin{abstract}
    Despite advances in upper-limb (UL) prosthetic design, achieving intuitive control of intermediate joints - such as the wrist and elbow - remains challenging, particularly for continuous and velocity-modulated movements. We introduce a novel movement-based control paradigm entitled Compensation Effect Amplification Control (CEAC) that leverages users' trunk flexion and extension as input for controlling prosthetic elbow velocity.

    Considering that the trunk can be both a functional and compensatory joint when performing upper-limb actions, CEAC amplifies the natural coupling between trunk and prosthesis while introducing a controlled delay that allows users to modulate both the position and velocity of the prosthetic joint.

    We evaluated CEAC in a generic drawing task performed by twelve able-bodied participants using a supernumerary prosthesis with an active elbow. Additionally a multiple-target-reaching task was performed by a subset of ten participants. Results demonstrate task performances comparable to those obtained with natural arm movements, even when gesture velocity or drawing size were varied, while maintaining ergonomic trunk postures. Analysis revealed that CEAC effectively restores joint coordinated action, distributes movement effort between trunk and elbow, enabling intuitive trajectory control without requiring extreme compensatory movements. Overall, CEAC offers a promising control strategy for intermediate joints of UL prostheses, particularly in tasks requiring continuous and precise coordination.
\end{abstract}


\section{Introduction}

Despite significant advances in mechanical design \cite{abayasiri2017mobio, resnik2014deka}, intuitive control of proximal intermediate joints—particularly the elbow and wrist—remains a critical challenge in (UL) prostheses. Although many contemporary devices allow for dexterous movements, most control paradigms struggle to reliably produce smooth, continuous motion at varying velocities, often forcing users into sequential joint motor actions and unnatural or compensatory postures.

EMG-based methods \cite{alshammary2017synergistic, farina2014extraction, fougner2012control} have proven robust and have been available since the 70s on commercial devices \cite{brack2021review, trent2020narrative}; however only a few propose velocity control (proportional to EMG amplitude) and only for hand gestures. Sequential control (SEQ) approaches typically use finite state machines to control one joint at a time with direct thresholded EMG control, while other methods employ pattern recognition \cite{connan2016assessmentEMG, jiang2012emg} or regression-based techniques \cite{hahne2014linear, smith2015evaluation, jiang2013accurate} to offer discrete control of a larger set of motor actions or even stiffness \cite{farina2024EMGImpedance}. Yet, these methods exhibit limited performance when precise, uninterrupted motion or coordinated joint movements are required. 

Recent movement-based control approaches offer a promising alternative by exploiting the motion of a specific body segment to drive one or several prosthetic joints, thereby establishing a synergy—the coordinated activation of multiple degrees of freedom in both the human and prosthetic limbs to achieve a task \cite{latash2007MotorSynergies,santello2013neural}. For instance, \cite{Merad2020} trained task-specific models from natural reaching data to predict elbow velocity from shoulder velocities during a pointing task, but did not solve the challenge of managing different learned synergies for multiple tasks, and of selecting the correct model for a given task in real time. This problem is addressed by integrating contextual target information—such as object position and orientation—into the prediction process \cite{Mick2021, Segas2023}, where prosthetic joints are driven by predictions from artificial neural networks that learn the mapping between the user’s residual limb (e.g. shoulder) kinematics and the distal joint angles. These approaches enable intuitive and adaptable control for reaching and grasping tasks, yet determining the target remains a large and complex challenge to be solved in a real use case scenario. Early intent-aware schemes already showed that driving a distal prosthetic joint from the leader’s end-effector velocity can lower shoulder and hip involvement during reaching tasks \cite{khoramshahi2021intent}.
%
\cite{GarciaRosas2020} presented a task-space synergy approach where the human–prosthetic interface leverages a kinematic model of the user’s arm and the anticipated path of the prosthetic hand rather than directly mapping residual limb joint movements to prosthetic joint movements. In another study, \cite{Haddadin2024} developed a synergy-based approach that uses low-dimensional residual limb movements to drive a 4-DoF prosthesis toward a specific goal. Although this method achieves continuous, velocity-based control in task space, its extension to real-time trajectory manipulation or general tracking tasks remains challenging, particularly since it relies on external motion capture systems and predefined endpoints. The Compensation Cancellation Control (CCC) approach proposed by \cite{Legrand2021} utilizes Acromion pose as the primary input for controlling elbow and wrist velocities. It operates, as under the assumption, as \cite{feder2024general}, that any deviation of an ideal upright posture is compensatory, and it adjusts the controlled joints to minimize or "cancel out" this deviation. Notably, CCC has already demonstrated effectiveness in a tracking task—such as tracing a wire loop—where continuous changes in goal positions and orientations are required \cite{legrand2020wireloop}. This success is partly due to the fact that wrist rotation can be controlled independently from the trunk's overall translational movements. In practice, only a minimal lateral trunk bend is needed to produce wrist rotation with CCC, and this slight movement does not affect the shoulder and legs' role in managing translation. However, tasks that demand a broader range of motion and a more integrated coordination between the prosthetic proximal joints and the user’s natural movements expose a key limitation: CCC assumes all trunk motion is purely compensatory. Because of this restrictive hypothesis, the controller struggles in scenarios where the trunk naturally serves a functional role and must actively contribute to reaching targets. Because CCC requires the trunk to pass back through the initial upright reference (i.e. non compensatory) posture to halt joint motion, every action becomes a two-phase maneuver: the user must first bend toward the target to initiate the prosthesis movement and then bend in the opposite direction to stop it. This interruption makes it extremely difficult to follow a continuous, precise hand trajectory.
\begin{figure*}[!h]
    \centering
    \includegraphics[width=0.75\textwidth]{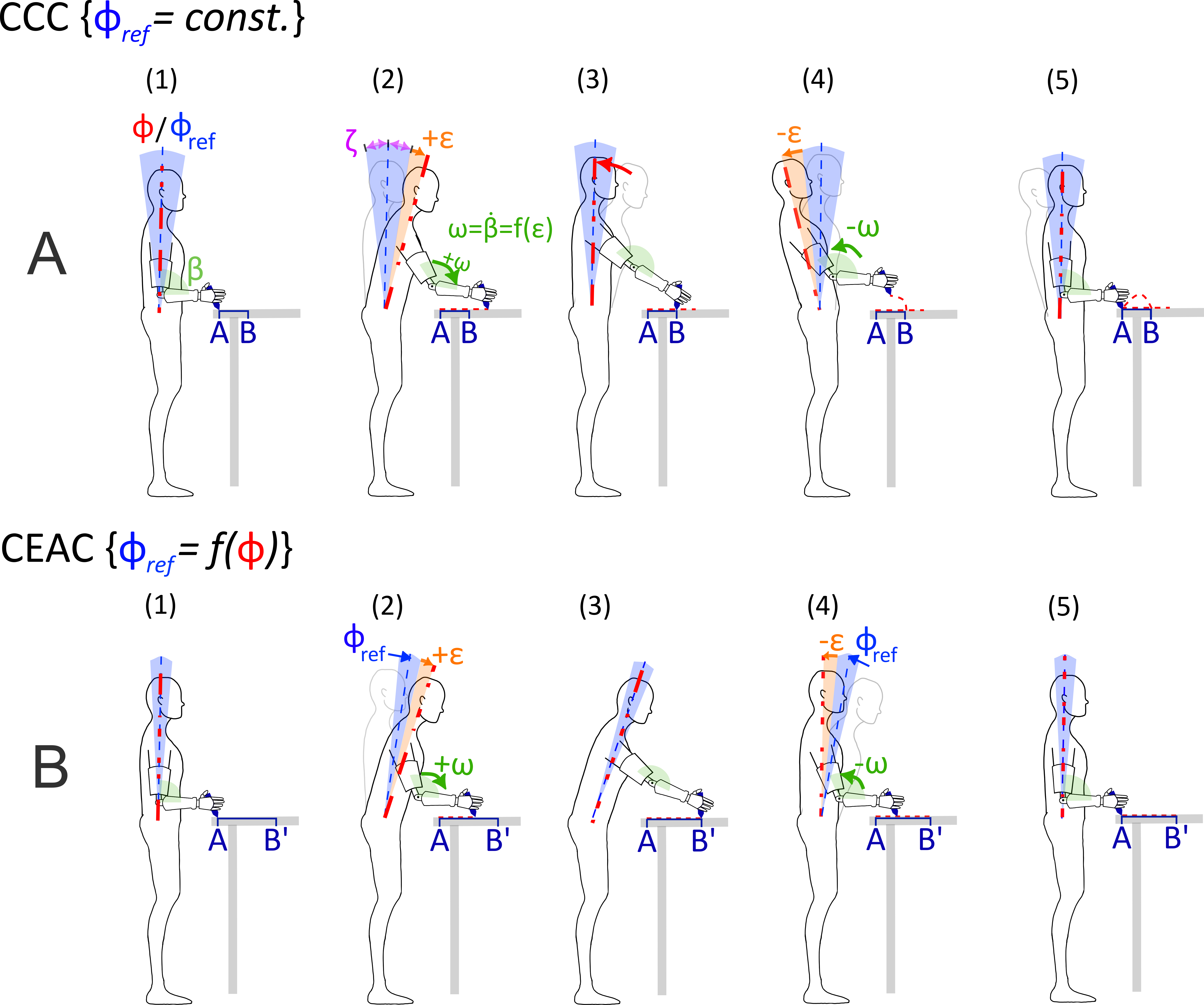}
    \caption{Illustration of how users of Compensation Cancellation Control (CCC) and Compensation Effect Amplification Control (CEAC) execute a round-trip line-drawing task. Points A and B denote the nominal start and end points of the path, while B' is an extended target. \textbf{Top row (A):} The sequence with CCC, where the trunk reference $\phi_{\text{ref}}$ is fixed.  \textbf{Bottom row (B):} The sequence with CEAC, where $\phi_{\text{ref}}$ is dynamic and follows the user's trunk with a delay. In both rows the movement is divided in five distinguished phases.}
    \label{fig::Figure_1_CCC_vs_CEAC_sketch}
\end{figure*}
While drawing a simple line on a table with an (UL) prosthesis may seem straightforward, it demands continuous, precise motion at varying speeds and close coordination between the user’s body and actively driven prosthetic joints. Existing control methods would struggle with such tasks: for instance, an EMG based strategy might prompt the user to lock the prosthesis in an optimal rigid posture and then use it as a rigid tool relying entirely on compensatory movements, leading to unergonomic body configurations and limited range of motion and accuracy \cite{metzger2012characterization}. Consequently, even basic drawing becomes a significant challenge for UL prosthetic users, making this scenario an interesting test case for evaluating movement control approaches.

To illustrate this, we consider CCC a strong candidate for such task, as it has already proven effective in a tracking task in the wire loop experiment, as mentioned earlier. 
Figure~\ref{fig::Figure_1_CCC_vs_CEAC_sketch}~(A) illustrates a typical challenge when using CCC to draw a simple forward-and-back line between points~A and~B(or B'). Initially upright~(1), the user bends the truk forward beyond a chosen deadzone $\zeta$~(2), generating a postural error that drives the prosthetic elbow toward the target. To stop the elbow motion precisely at B, however, the user must lean backward, overshooting the target posture due to the required return through the upright reference~(3). Additionally, reversing the movement from B to A necessitates ergonomically problematic backward trunk bending~(4), causing the pen tip to lift off the surface unless complicated compensations involving hip and leg adjustments are made. This two-phase trunk motion severely limits CCC’s effectiveness for continuous and precise trajectory tasks.

To address these critical limitations, we introduce Compensation Effect Amplification Control (CEAC), a novel movement-based strategy featuring a dynamically adapting reference posture instead of CCC’s fixed (here upright) reference position. By adjusting the reference to follow the trunk’s posture with a controlled delay, we hypothesise CEAC could enable an intuitive velocity-based control of multiple prosthetic joints, utilizing trunk motion as a functional rather than purely compensatory input. We begin by presenting the core principles of CEAC and illustrate their functional implications through a targeted demonstration of a forward-and-backward line-drawing task, recorded in an illustrative trial with a trained able-bodied user. Subsequently, we experimentally evaluate its performance with able-bodied participants (wearing a supernumerary UL prosthesis) on both a continuous drawing task and a multi-target pointing task. Finally, we present quantitative results and discuss their significance.

\section{Controller principles}

Unlike its predecessor CCC, which treats trunk motion solely as a compensatory action to be canceled, CEAC leverages trunk flexion as a mostly functional control input. By using a dynamic reference posture that follows the actual posture with a delay, the controller amplifies the natural coupling between trunk and elbow movements while introducing a controlled delay that allows the user to modulate both position and velocity of the prosthetic elbow. 

The CEAC paradigm is designed to operate within the shared kinematics of the human-prosthesis system. In this integrated model, the velocity of the end-effector (i.e. the prosthetic hand), $\dot{\mathbf{x}}_e$, is determined by the combined contributions of the user's biological joints ($\dot{\mathbf{q}}_h$) and the prosthetic joints ($\dot{\mathbf{q}}_p$):
\begin{equation}
    \dot{\mathbf{x}}_e = \mathbf{J}_{he}\,\dot{\mathbf{q}}_h + \mathbf{J}_{pe}\,\dot{\mathbf{q}}_p,
\end{equation}
where $\mathbf{J}_{he}$ and $\mathbf{J}_{pe}$ are the respective Jacobian matrices. In the general case, compensatory movements can be described by tracking the state of a "compensatory frame," $\mathbf{x}_c$, whose velocity is related to the user's joint velocities by its own Jacobian, $\mathbf{J}_c$, such that $\dot{\mathbf{x}}_c = \mathbf{J}_c(\mathbf{q_h},\mathbf{q_p})\,\dot{\mathbf{q}}_h$ (for $\dot{\mathbf{q}}_p=0$), as described in \cite{feder2024general}.

\subsection{From General Kinematics to a Simplified Model}
A controller based on the general kinematic model would need to translate a compensatory error into a desired motion for the prosthetic joints ($\dot{\mathbf{q}}_p$). This would typically require solving a complex inverse kinematics problem, which is computationally expensive and sensitive to modeling errors and kinematic singularities. 

For CEAC, we make a deliberate design choice to bypass this complexity in favor of a more robust and intuitive approach. This simplification involves two key steps:
\begin{enumerate}
    \item \textbf{Input Simplification:} Based on empirical observations of forward-reaching tasks, we identify that trunk flexion-extension is the dominant component of the compensatory motion. We therefore use the trunk angle, $\phi(t)$, as a high-fidelity, one-dimensional proxy for the entire multi-dimensional compensatory state $\mathbf{x}_c$. This allows for a simple and robust input signal that can be measured with a single wearable sensor, avoiding the need for a full-body kinematic model.
    
    \item \textbf{Direct Functional Mapping:} Instead of using this input to calculate a target for the end-effector and then solving the prosthesis's inverse kinematics (which would involve $\mathbf{J}_{pe}$), we establish a new, direct functional synergy. We directly map the trunk motion input ($\phi(t)$) to the prosthetic elbow velocity ($\omega_{\text{elbow}}(t)$). Yet this is not an attempt to mimic natural human synergies but to design an effective and intuitive flexible human-machine synergy. This approach deliberately simplifies the mapping to an identity relationship, where the "prosthesis error" is considered to be directly represented by the compensatory error, making the control law computationally light and transparent.
\end{enumerate}

\subsection{Expected behavior}
Before delving into the controller specifics, we revisit the Fig.~\ref{fig::Figure_1_CCC_vs_CEAC_sketch} (B), which illustrates the same line-drawing task but now employing a dynamic reference posture following the current posture with a delay. Initially (1), the user stands upright with the elbow flexed, mirroring the starting condition in CCC. As soon as the user leans forward beyond the deadzone $\zeta$(2), the elbow extends to assists the motion in the direction of trunk bending. Subsequently, the user may either i) lean back slightly to catch the deadzone (which is converging toward current posture) and align their current posture with the new reference posture zeroing the error and thereby stopping the elbow motion or ii) cease trunk motion at just the right time and letting the reference catch the current posture slightly after, allowing the elbow to complete its movement while the shoulder continues to adjust. In state~(3), the pen is at point~B' (further away than B) with the trunk still inclined forward. Crucially, this dynamic reference means there is no need to return to an upright posture to halt the elbow—thereby avoiding the necessity of an overshoot—and the trunk can remain in a more functional orientation for the task, thus extending the size of the workspace. Once the reference posture $\phi_{\text{ref}}$ aligns with the user’s current trunk angle $\phi$, the trunk is effectively locked in at a new forward-lean position (3). From here, the user can bend backward to generate a negative error, reversing the elbow motion and ultimately returning to the initial state (5) without the need of trunk over-extension like in the CCC example in (4).

Overall, we hypothesize that this dynamic reference approach should allow the trunk to still act as a truly functional input, not only extending the controllable range of motion but also eliminating the previous necessity of unergonomic backward trunk bending. Most importantly, it should prevent the two-phase trunk motion characteristic of previous strategies, in which the user, after initially leaning toward the target, must move in the opposite direction to halt the prosthetic movement. Avoiding this counter-intuitive requirement should greatly enhances control intuitiveness and precision. Additionally, because the reference posture $\phi_{\text{ref}}$ continually updates to follow the current trunk angle $\phi$, users should be able to realign more easily with the reference. Consequently, a smaller deadzone could be selected,resulting in higher responsiveness of the elbow joint to subtle movements, without compromising precise user control.

\subsection{Controller Fundamentals}

The CEAC controller computes a dynamic reference posture $\phi_{\text{ref}}(t)$ by applying a first-order low-pass filter to the actual trunk angle $\phi(t)$, with cutoff frequency $f_c$:

\begin{equation}
\label{eqn:ref_definition}
\phi_{\text{ref}}(t) + \tau\,\frac{d\phi_{\text{ref}}(t)}{dt} = \phi(t),
\end{equation}

where $\tau = \frac{1}{2\pi\,f_c}$ is the filter’s time constant. 

The error signal driving the elbow velocity command is defined as the difference between the current trunk posture and this dynamic reference:

\begin{equation}
\epsilon(t) = \phi(t) - \phi_{\text{ref}}(t).
\end{equation}

Elbow velocity is controlled proportionally based on this error, incorporating a deadzone $\zeta$ to avoid unintended elbow motion due to minor trunk fluctuations:

\begin{equation}
\omega_{\text{elbow}}(t) = 
\begin{cases}
0, & \text{if } |\epsilon(t)| \leq \zeta,\\[6pt]
\lambda \Bigl(\epsilon(t)-\zeta\,\text{sign}(\epsilon(t))\Bigr), & \text{if } |\epsilon(t)| > \zeta,
\end{cases}
\end{equation}

where $\lambda$ is a gain factor. No saturation is used as we consider that the elbow velocity $\omega_{\text{elbow}}(t)$ is generally capped by the actuator's maximum feasible velocity.

To minimize unergonomic backward trunk bending, the dynamic reference update is paused (effectively, $f_c = 0$) whenever the reference angle $\phi_{\text{ref}}$ surpasses the initial upright orientation $\phi_{\text{ref},0}$. 
\begin{equation}
f_c(\phi_\mathrm{ref}) =
\begin{cases}
f_{c}, &  \text{if } \phi_{\mathrm{ref}}(t) \ge \phi_{\mathrm{ref},0},\\[6pt]
0, & \text{if } \phi_{\mathrm{ref}}(t) < \phi_{\mathrm{ref},0},
\end{cases}
\label{eq:freeze_ref}
\end{equation}

This assumes backward trunk bending is typically compensatory rather than functional.

\subsection{Conceptual differences from synergy control}
\label{subsec:conceptual_difference}

\paragraph*{1) Frequency–domain view}
The low-pass relationship defining the reference angle $\phi_{\text{ref}}$ is expressed in the frequency domain as:

\begin{equation}
    \Phi_{\text{ref}}(s) = \frac{\Phi(s)}{\tau s + 1},
\end{equation}

where $\Phi_{\text{ref}}(s)$ and $\Phi(s)$ represent the Laplace transforms of the reference and trunk angles, respectively.

The error $E(s)$ driving elbow velocity is therefore:

\begin{equation}
    E(s) = \Phi(s) - \Phi_{\text{ref}}(s) 
         = \Phi(s) - \frac{\Phi(s)}{\tau s + 1} 
         = \frac{\tau\,s}{\tau s + 1}\Phi(s).
\end{equation}

In the simplified linear case ($\zeta=0$), the CEAC law gives the transfer function:
\[
    \frac{\Omega_{\text{elbow}}(s)}{\Phi(s)}
        = \lambda\frac{\tau s}{1+\tau s},
\]
i.e.\ a first-order \emph{high-pass} between trunk \emph{position} $\Phi(s)$ and elbow \emph{velocity} $\Omega_{\text{elbow}}(s)$.
\begin{figure*}[!ht]
    \centering
    \includegraphics[width=0.98\textwidth]{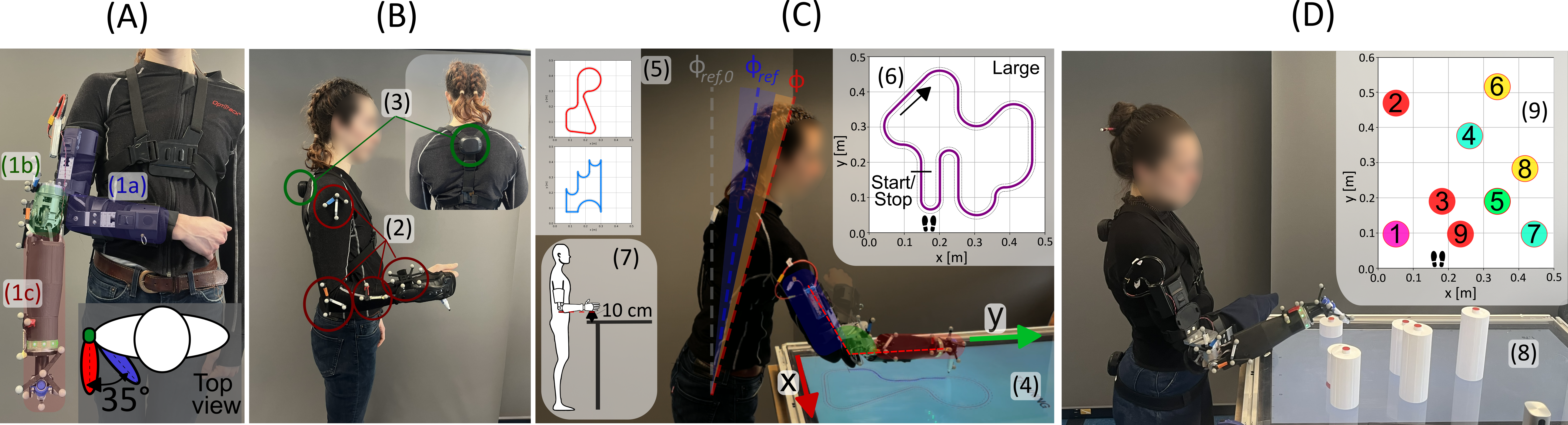}
    \caption{Overview of the experimental setup. 
    \textbf{(A)}~Prosthesis configuration with an active elbow (1b), a prosthetic forearm (1c), and a blocking elbow orthosis (1a) attached to the natural limb. The prosthetic forearm is oriented at $35^\circ$ relatively to the natural forearm. OptiTrack markers are placed on the pen, the prosthetic forearm, above the elbow, and on the upper arm. 
    \textbf{(B)}~Natural-limb setup with markers (2) on the upper arm, elbow, and forearm, as well as a wrist orthosis to which the pen is rigidly attached. In both setups, a VIVE\textregistered\ Ultimate Tracker is mounted above the trapezius (3). 
    \textbf{(C)}~A participant performing the drawing task on a Plexiglas-covered screen (4). The top left of panel C shows two training paths (5), while the main experimental path is shown enlarged in the top right (6). The bottom left illustrates the table height relative to the participant (7). 
    \textbf{(D)}~A participant performing the reaching task using a Plexiglas-covered screen and 3D-printed cylinders as targets (8). The corresponding target positions are shown in screen coordinates (9), with numbers indicating the sequence in which the targets are to be reached. Targets 2, 3, and 9 are flush with the screen surface (0\,cm depth), target 1 is placed at 5\,cm, target 5 at 10\,cm, targets 4 and 7 at 15\,cm, and targets 6 and 8 at 20\,cm from the screen.}
    \label{fig::Figure_2_ExperimentalSetup}
\end{figure*}

\paragraph*{2) Integrated (position–position) effect}
Integrating elbow velocity corresponds to multiplying by $1/s$ in Laplace space:
\[
    \Theta_{\text{elbow}}(s)
      = \frac{1}{s}\,\Omega_{\text{elbow}}(s)
      = \lambda \frac{\tau}{1+\tau s}\,\Phi(s).
\]
For a finite trunk step $\Delta\phi$ (\(\Phi(s)=\Delta\phi/s\)) this yields
\[
    \Theta_{\text{elbow}}(s)
      = \lambda \frac{\tau}{s(1+\tau s)}\Delta\phi
      \;\;\Longrightarrow\;\;
    \Delta\theta = \lambda\tau\,\Delta\phi,
\]
showing that the \emph{final} elbow-angle change depends only on the net trunk excursion, irrespective of movement speed.  
The same result appears directly in the time domain.  Integrating the low-pass definition
\(
 \phi_{\text{ref}}+\tau\dot{\phi}_{\text{ref}}=\phi
\)
from \(t=0\) (trunk angle $\phi_{0}$) to \(t=\infty\) (trunk angle $\phi_{0}+\Delta\phi\)) gives
\[
\boxed{\;
  \int_{0}^{\infty}\bigl[\phi(t)-\phi_{\text{ref}}(t)\bigr]dt
      = \tau\,\Delta\phi
\;}
\]
and because CEAC commands
\(
 \omega_{\text{elbow}} = \lambda\, \bigl[\phi-\phi_{\text{ref}}\bigr],
\)
integration over time again produces
\(
 \Delta\theta = \lambda\tau\Delta\phi.
\)

\paragraph*{3) Transient lag and velocity modulation (“release-catch”).}
Although steady-state behaviour is position-proportional, the reference $\phi_{\text{ref}}$ is generated by a first-order \emph{low-pass}, so it lags the real trunk posture by a time constant $\tau$.  
A rapid trunk flexion that moves the posture far ahead of the lagging reference creates a large positive error and therefore drives elbow extension; as soon as the user partially reverses the trunk motion while the reference is still advancing, the error diminishes toward zero and the elbow velocity falls to zero, effectively stopping the prosthetic joint.  
Alternating such brief “forward/back” micro-cycles lets the user sculpt continuous elbow velocity without altering the ultimate relationship \(\Delta\theta =\lambda\tau\Delta\phi\). 

\paragraph*{4) No-effect trunk velocity (deadzone).}
With a finite deadzone \(\zeta>0\) the elbow remains still whenever
\[
  |\phi(t)-\phi_{\text{ref}}(t)|\le\zeta .
\]
Using the low-pass relation yields the maximum trunk speed that \emph{does not} move the elbow:
\[
  |\dot{\phi}_{0,\max}| = 2\pi f_c\,\zeta .
\]
The deadzone ($\zeta = 2^{\circ}$) was set just above the amplitude of natural trunk sway, preventing spurious activation while still responding to intentional flexion.
The cut-off frequency ($f_c = 0.1\;\text{Hz}$; $\tau \approx 1.6\;\text{s}$) was chosen empirically as the smallest value that produced a perceptible but brief lag, keeping the system responsive while giving the user enough time to modulate elbow motion.
Moving their trunk at a speed below $|\dot{\phi}_{0,\max}|$ allow users to reposition their trunk without actuating the elbow, whereas faster motions engage elbow velocity control.

\paragraph*{5) Summary}
While conventional synergy controllers implement an instantaneous velocity-to-velocity mapping $S$ between joints,
\(
 \dot q_{\text{device}} = S\dot q_{\text{body}},
\)
CEAC instead maps a \emph{lagged high-pass} of trunk position to elbow velocity,
\[
  \dot q_{\text{device}}=\lambda\,\frac{\tau s}{1+\tau s}\,q_{\text{body}},
\]
providing: 
\begin{enumerate}
    \item[i)] immunity to slow postural drift,  
    \item[ii)] intuitive velocity modulation via trunk “release-catch”, 
    \item[iii)] in the idealised linear case ($\zeta = 0$) the final elbow excursion is proportional to the net trunk excursion
      ($\Delta\theta = \lambda\tau\,\Delta\phi$);
\item[iv)]in the practical non-linear case ($\zeta>0$) this proportionality
      applies only to the part of the trunk motion that (a) lies outside the
      deadzone and (b) exceeds the “no-effect’’ speed
      $|\dot\phi_{0,\max}|=2\pi f_c\zeta$  movements that stay within the deadzone or below that speed. Otherwise those would leave the elbow unchanged, giving the user additional freedom to fine-tune both elbow velocity and final angle.

\end{enumerate}

\section{Materials and Methods}

\subsection{Experimental setup}

Figure~\ref{fig::Figure_2_ExperimentalSetup} provides an overview of the experimental setup. Panel~(A) illustrates the \textit{prosthesis} configuration applied to an able-bodied participant. In this setup, a supernumerary prosthesis is attached to the natural limb via a blocked-elbow orthosis (1a) to ensure stable fixation. The prosthesis features an active elbow (1b) and a prosthetic forearm (1c) oriented at approximately \(35^\circ\) relative to the natural forearm fixation. Optical marker clusters are mounted on the pen, the prosthetic forearm, the prosthetic elbow, and above the prosthetic elbow.

Panel~(B) depicts the \textit{natural-limb} configuration, in which the participant’s upper arm, elbow, and forearm are each fitted with optical marker clusters, and the wrist is immobilized with a wrist-blocking orthosis to which the pen is rigidly attached. The wrist is blocked to be able to better compare the natural limb with the supernumerary prosthesis which does not possess an articulated wrist. In both setups, a VIVE Ultimate Tracker\textregistered\ is positioned above the trapezius, centered between the shoulder blades, to measure trunk flexion. This uniform approach to measuring trunk (with VIVE Ultimate Tracker) and elbow (with MoCap) angles enables direct comparisons between the prosthesis and natural-limb conditions. It is important to note that the OptiTrack system is used solely for recording movements during the experiment and does not serve as an input to our controller. In contrast, the prosthesis relies exclusively on a VIVE tracker—a visual-inertial odometry device that employs SLAM algorithms via inside-out tracking.
\textit{UL Prosthesis:} The prosthesis is powered by a 2S LiPo battery and controlled by a Raspberry~Pi with a custom shield. The active elbow (Fillauer\textsuperscript{\textregistered} Hosmer E-TWO Electric Elbow) is capable of flexing against gravity at speeds up to approximately 50\,deg/s. The whole prototype with fixation and battery weighs 1.6kg.\\
\textit{VIVE tracker:} Data from the VIVE tracker is processed by a custom Python application~\cite{MyRepo} that communicates with the prosthesis via UDP, while the prosthesis communicates with the main experiment laptop via MQTT. This laptop runs a custom Pygame-based drawing interface and logs all OptiTrack data—tracking both the elbow angle and pen position—synchronously at 60\,Hz. In parallel, trunk orientation is recorded at 60\,Hz via the VIVE tracker. This unified data pipeline ensures that all kinematic information is captured synchronously.\\
\textit{Motion capture:} Five OptiTrack cameras (two PrimeX~13W and three Prime~13) are mounted at a height of approximately 2.8\,m around the workspace, each positioned roughly 3.5\,m from the drawing screen.\\
\textit{Drawing table:} The table, which supports a 42-inch screen covered by a rigid Plexiglas layer, is adjustable in height so that it can be positioned 10\,cm below the participant’s elbow when the shoulder is neutrally aligned with the torso (Fig.~\ref{fig::Figure_2_ExperimentalSetup},C,7).\\
\textit{CEAC tuning:} The gains were tuned through preliminary testing and kept constant across all participants. In our implementation, we set \(\zeta = 2^\circ\), \(f_c = 0.1\,\mathrm{Hz}\), and \(\lambda = 3\,\mathrm{s}^{-1}\); these values ensured that the user maintains robust controllability while still experiencing a responsive, natural feel.

\subsection{Protocol}

The experimental procedure consisted of several distinct phases.  


\paragraph*{Preliminary validation trials}
Prior to conducting the main experimental tasks, two preliminary demonstrations were carried out to qualitatively validate CEAC’s hypothesised behaviour as illustrated in Fig.,\ref{fig::Figure_1_CCC_vs_CEAC_sketch}: a single-user comparison between CCC and CEAC in a simple line-drawing task, and a set of trials where one trained participant executed the line-drawing task at five different speeds using the CEAC-controlled prosthesis.

\paragraph*{Drawing task}  
Drawing data were recorded virtually using a custom Python/PyGame program, while the pen remained ink-free.  A trace was rendered whenever the pen tip approached within 2 mm of the Plexiglas-covered display.  During the training phase (that lasted for each participant 10 to 15 min for the drawing task and 5min for the reaching task), participants practised drawing basic paths (squares, circles, ellipses) as well as two racetrack-like contours (Fig.\ref{fig::Figure_2_ExperimentalSetup} C.5).  Each training path was circumscribed by a 1 cm tolerance zone that provided immediate visual feedback—the screen background turned green if the pen tip remained within the zone and white otherwise.\\
Following training, participants were instructed to draw a novel path (Fig.\ref{fig::Figure_2_ExperimentalSetup} C.6) that was not used during training, thereby enabling evaluation of their generic drawing ability.  This path, characterised by multiple turns and direction changes, required continuous velocity modulation. Three speed instructions (\emph{SLOW}, \emph{MEDIUM}, \emph{FAST}) were displayed above the path, and the same path was presented in two sizes (approx.\ 0.40 m × 0.40 m, \emph{large}, and 0.28 m × 0.28 m, \emph{small}). For the \emph{MEDIUM} speed condition, participants were instructed to maintain a comfortable pace that allowed them to consistently remain within the 1 cm tolerance zone. For the \emph{SLOW} condition, the emphasis was on maximizing precision by moving at a deliberately reduced speed. Conversely, for the \emph{FAST} condition, participants were asked to maximize their speed, even if it meant a compromise in precision. Each speed–size combination was repeated four times, yielding 24 trials per participant (6 conditions × 4 repetitions).  The participant’s stance relative to the path is illustrated qualitatively in Fig.\ref{fig::Figure_2_ExperimentalSetup}, C.6.\\
The drawing experiment was conducted under two arm conditions: once with the supernumerary prosthesis and once with the natural limb equipped with a wrist-blocking orthosis.  The order of arm conditions was alternated between participants.  Within each session, the drawing block followed a fixed internal sequence: large path at all three speeds, then small path at all three speeds.

\paragraph*{Reaching task}  
Upon completion of the drawing block, participants moved on to a three-dimensional reaching task (Fig.\ref{fig::Figure_2_ExperimentalSetup}, D).  
Each round involved nine cylindrical targets presented in a fixed sequence.  
The pen tip had to enter a 1 cm‐radius tolerance sphere centred on the active target and remain there for 2s, after which the next target was cued; visual feedback signalled entry into the sphere and the countdown. 
Thus a round comprised nine transport movements interleaved with nine 2s dwell periods.\\
Participants first completed one to two practice rounds with an alternative target positioning and order, then performed three recorded rounds using the predefined sequence.  
The initial arm condition (natural elbow or prosthesis) alternated between consecutive participants.\\
No constraints were imposed on trunk or shoulder motion; participants were free to adopt any posture that facilitated target acquisition. The foot placement, however was fixed: marks on the floor indicated where to stand, and the participants were instructed not to move their feet during the task.

\subsection{Participants}

Twelve able-bodied participants (6 male, 6 female) took part in the study.  
All were free of musculoskeletal or neurological impairments, and ranged in age from 22 to 33 years and in stature from 170 to 196 cm.  
Every participant completed the drawing-task protocol, while a subset of ten (5 male, 5 female) also performed the nine-target-reaching task; the remaining two could not attend the additional session for scheduling reasons.  
Each participant provided written informed consent prior to testing.  

\subsection{Data processing and Statistical Analysis}
\label{StatisticalAnalysis}

All angular and positional time series were zero-lag filtered with a third-order Butterworth filter (\textit{filtfilt} implementation, 5 Hz cut-off).  
The 5~Hz bandwidth captures the upper end of voluntary upper-limb and trunk motion motion while removing measurement noise; the zero-phase (forward–reverse) application prevents temporal shifts that could bias velocity-based metrics.

All statistical tests were conducted in Python using SciPy. Differences between natural elbow and prosthesis were assessed via independent two-sample t-tests (unequal variance assumed) for each combination of intended speed (\emph{MEDIUM}, \emph{SLOW}, and \emph{FAST}) and path size (Large, Small). 

Significance levels were defined as: \textit{ns} ($p > 0.05$), * ($p \leq 0.05$), ** ($p \leq 0.01$), *** ($p \leq 0.001$). Repetition outliers beyond three standard deviations from the mean were excluded within each condition group.

\subsection{Assessment metrics}
\label{subsec:assessment_metrics}

For each trial, a consistent set of task-level and joint-level metrics was extracted.

\paragraph{Task-level metrics}
\begin{itemize}
  \item \textbf{Completion time} — for the drawing task, the time elapsed from the first pen displacement beyond the start line to the final return and immobility at that line; for the reaching task, time from the appearance of the first target until the pen tip made stable contact with the ninth (final) target.
  \item \textbf{Precision score} (drawing task) — mean Euclidean distance between the recorded pen-tip trajectory and the reference path, sampled every 0.5 mm along the drawn curve.
  \item \textbf{Path-length ratio (PLR)} (reaching task)  
        \[
        \text{PLR}=\frac{\displaystyle\sum_{i=1}^{N-1}\lVert\mathbf{p}_{i+1}-\mathbf{p}_{i}\rVert}
                          {\lVert\mathbf{p}_{N}-\mathbf{p}_{1}\rVert},
        \]
        where \(\mathbf{p}_{i}\) is the pen-tip position at sample \(i\).  
        A value of 1 indicates an ideal straight-line trajectory.
  \item \textbf{Spectral arc length (SPARC)} — smoothness index of the tangential-speed profile \cite{balasubramanian2015analysis}.  
        After zero-meaning and unit-energy normalization of \(v(t)\), the negative arc length of the log-magnitude FFT between 0.1 rad s\(^{-1}\) and the –40 dB cut-off is computed; less-negative values correspond to smoother movements.
\end{itemize}

\paragraph{Joint-level metrics}
\begin{itemize}
  \item \textbf{Range of motion (ROM)} — difference between the maximum and minimum joint angles recorded in the trial.
  \item \textbf{Total amount of movement} — the cumulative angular change that reflects how much a joint “travels,” irrespective of direction,
        \[
        \text{Total movement} = \sum_{i=1}^{N-1} \bigl|\,\theta_{i+1}-\theta_{i}\bigr|,
        \]
        with \(\theta_{i}\) the joint angle at time step \(i\).
  \item \textbf{Synergy Joint Velocity Index (SJVI)} — a measure of simultaneous joint activation.  
        For each inter-target transport phase the absolute angular velocities of elbow and shoulder are obtained, excluding the final 2 s of that phase (during which the target is being held).  
        At every remaining sample the velocities are classified as “active’’ when \(|\dot{\theta}| \ge 1^{\circ}\,{\rm s}^{-1}\).  
        The SJVI for that phase is the proportion of samples in which \emph{both} joints are active:  
        \[
        \text{SJVI}_{k}=\frac{\text{\,\#\,samples with }(\dot{\theta}_{\text{el}}\ge1^{\circ}\!/{\rm s}\;\land\;\dot{\theta}_{\text{sh}}\ge1^{\circ}\!/{\rm s})}
                              {\text{total samples in phase }k}.
        \]
        The trial-level SJVI is the mean of the eight phase values, yielding 0 when the joints never move together and 1 when they are always co-active.
\end{itemize}

\section{Results}

\textit{The accompanying multimedia attachment contains video recordings of representative trials from a subset of the experimental tasks described in this section.}

\subsection{Preliminary experiments}

\subsubsection{Drawing a line - qualitative comparison}
\label{subsec:qual_example_line}

\begin{figure}[!ht]
    \centering
    \includegraphics[width=0.45\textwidth]{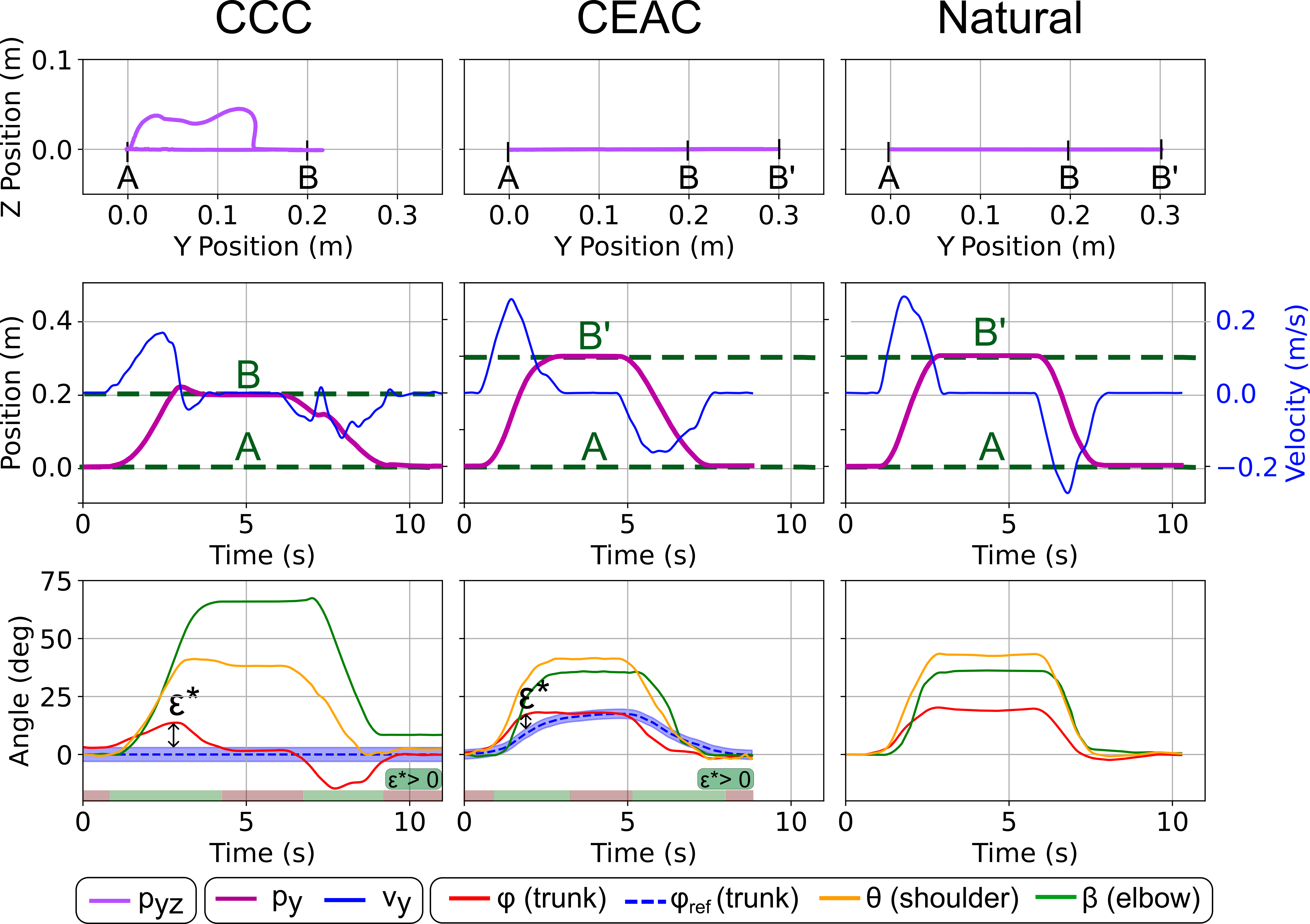}
    \caption{Drawing a 20\,cm line with three control conditions. Columns compare Compensation Cancellation Control (CCC), Compensation Effect Amplification Control (CEAC), and the natural elbow. In the plots, labels A, B, and B' denote the start, nominal end, and extended/overshoot points along the path. Rows show: pen-tip trajectory in the $yz$-plane (top), pen-tip position and velocity over time (middle), and joint angles over time (bottom). The horizontal ribbon indicates when the elbow motor is active ($\epsilon^{*}>0$).}
    \label{fig::Figure_3_CCCvsCEACvsNatural_exp}
\end{figure}

Figure~\ref{fig::Figure_3_CCCvsCEACvsNatural_exp} compares a forward-and-back line drawing (20 cm) performed using Compensation Cancellation Control (CCC), Compensation Effect Amplification Control (CEAC), and the participant’s natural limb with a blocked wrist. To ease the comparison between trunk, shoulder and elbow trajectories, all angles are expressed relative to the initial posture, where 0$^{\circ}$ corresponds to the starting angle of the trunk and elbow.

In the CCC condition (left column), the lower panel reveals that, to halt elbow extension at point B, the participant must pass the trunk posture back through the upright reference, reducing the postural error to zero. This obligatory trunk reversal forces the participant to overshoot the target at the end-effector level, clearly observable in the upper and middle panels at position B. Additionally, returning from B to A necessitates trunk bending in the opposite direction under CCC, which results in the pen tip lifting off the drawing surface. Moreover, initiating elbow flexion during the return stroke requires ergonomically disadvantageous backward trunk bending.

Conversely, in the CEAC condition (center column), the trunk acts functional: once the dynamic reference posture has moved forward, the participant can precisely halt elbow movement at the target point without reverting to an upright trunk posture. This allows participants to comfortably maintain a functional trunk oriented toward the target, effectively reaching further position B'. The pen remains consistently in contact with the drawing surface, and joint-angle profiles closely resemble those of natural-limb movement. The horizontal ribbon beneath each plot (green = elbow motor active, red = inactive) clearly demonstrates continuous elbow actuation whenever the pen is in motion under CEAC.

Comparing CEAC to natural-limb control (right column) indicates highly similar kinematic strategies across joints. Overall, the CCC condition distinctly differs, resulting in altered trunk posture patterns, excessive elbow use, and reduced workspace, while CEAC closely mirrors the natural joint kinematics.

\subsubsection{Speed modulation with fixed gains}
\label{subsec:qual_speed_mod}

\begin{figure*}[!ht]
    \centering
    \includegraphics[width=0.8\textwidth]{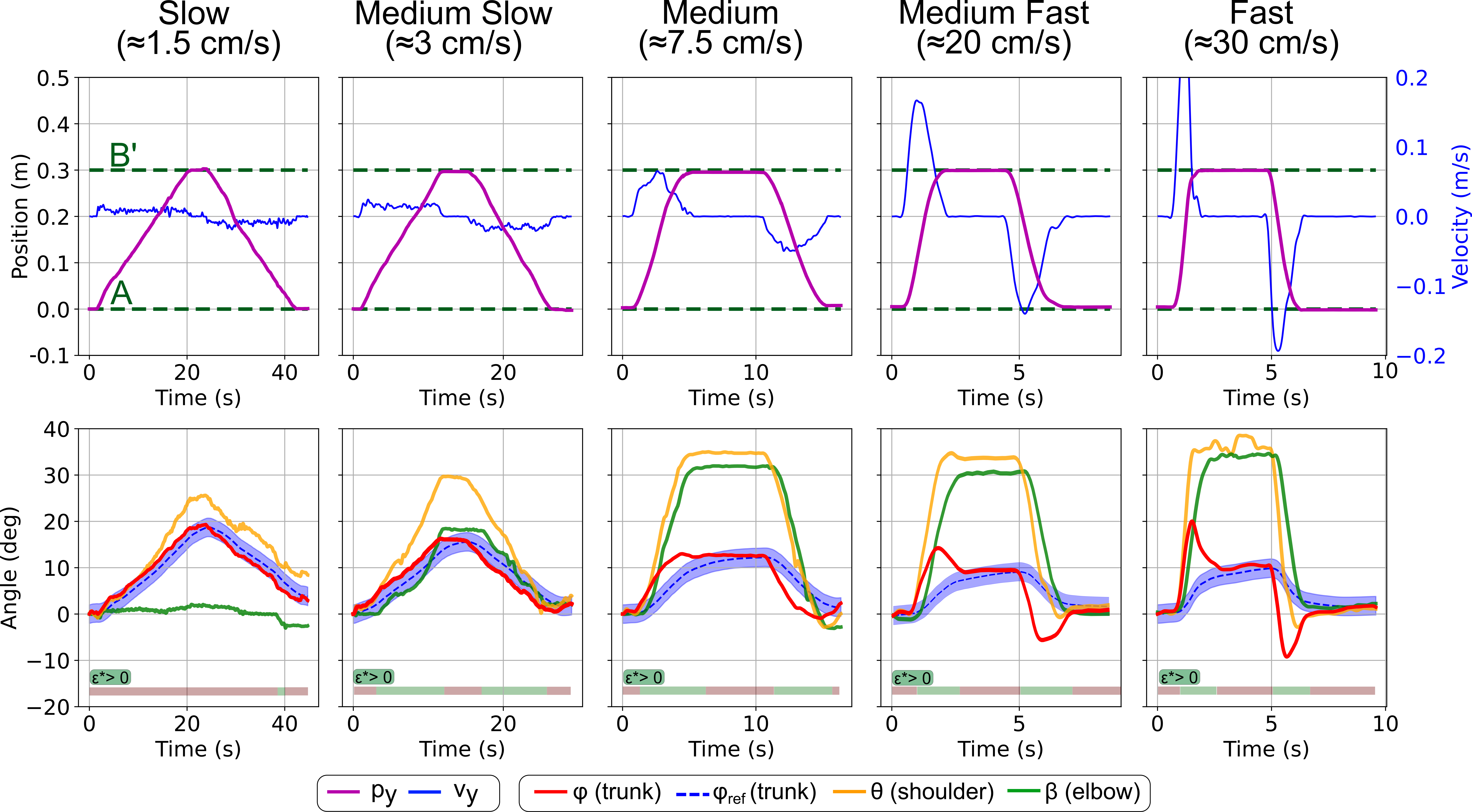}
    \caption{Pen-tip and joint kinematics during line-drawing trials with the CEAC-controlled prosthesis at five different instructed speeds. Top row: pen-tip position ($p_y$) and velocity ($v_y$), where dashed lines label the start (A) and end (B') positions. Bottom row: trunk ($\phi$), shoulder ($\theta$), and elbow ($\beta$) joint angles. The horizontal ribbon indicates periods of active elbow control ($\epsilon^{*}>0$).}
    \label{fig::Figure_4_CEAC_5speeds_exp}
\end{figure*}

Figure~\ref{fig::Figure_4_CEAC_5speeds_exp} demonstrates how a trained participant modulates line-drawing speed using consistent CEAC controller gains across trials. The top-row panels illustrate variations in peak pen-tip velocities, ranging from slow (approximately 1.5 cm/s) to fast (approximately 30 cm/s), yet consistently maintaining a coherent velocity profile and continuous contact with the drawing surface. To ease the comparison between trunk shoulder and elbow trajectories, all angles are expressed relative to the initial posture, where 0$^{\circ}$ corresponds to the starting angle of the trunk and elbow. 

In the lower-row panels, it is evident that the participant achieves elbow velocity modulation primarily by altering the amplitude and timing of short trunk "release-catch" motions, capitalizing on the lag of the dynamic reference posture $\phi_{ref}$. 
Notably, the extreme case depicted in the leftmost column shows such slow trunk movement that trunk position remains entirely within the deadzone, and trunk velocity never exceeds the $|\dot\phi_{0,\max}|$ threshold. Consequently, the elbow remains inactive, and the task is executed purely through compensatory movements of the trunk and shoulder.

\subsection{Path tracking}

\subsubsection{Illustrative example of the Drawing Task}
\begin{figure}[!ht]
    \centering
    \includegraphics[width=0.5\textwidth]{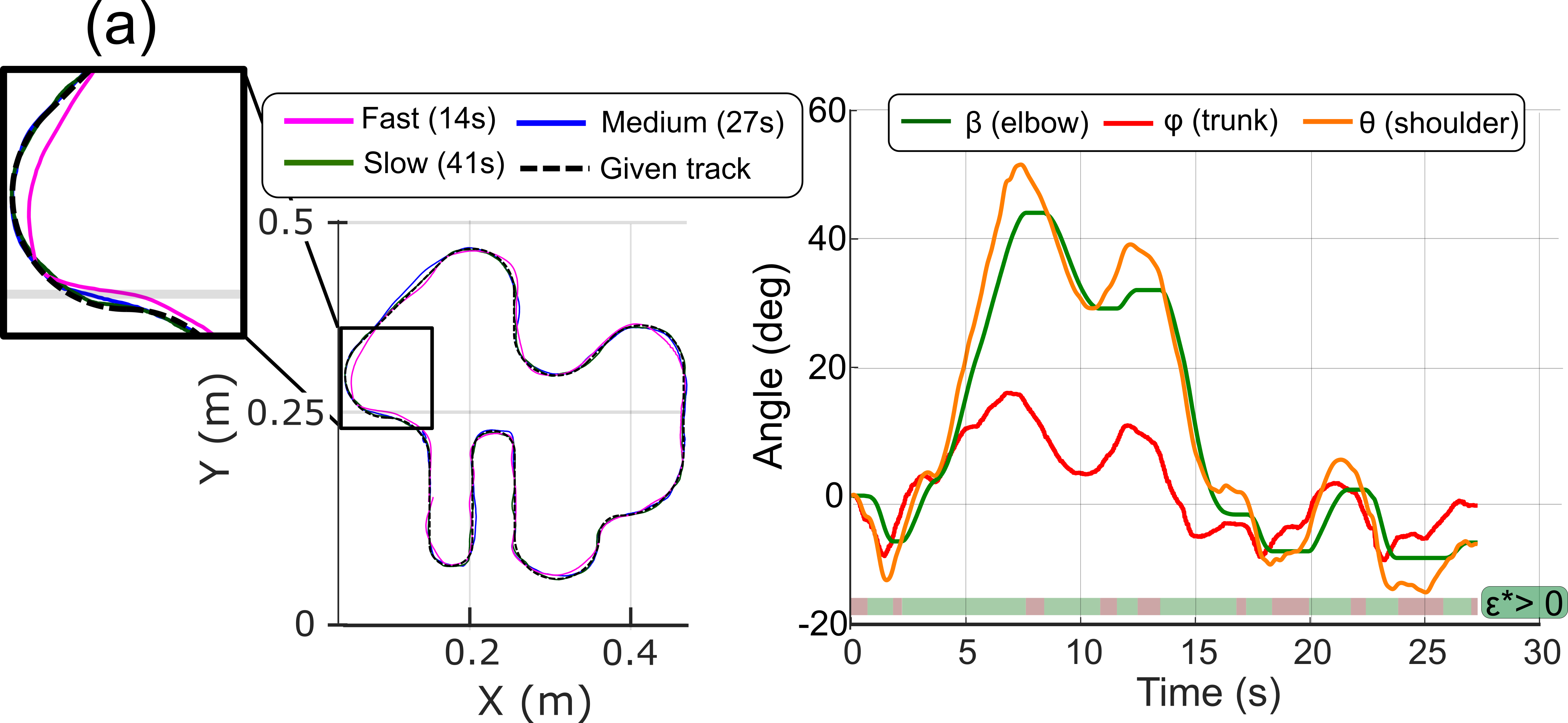}
    \caption{Qualitative example of drawing the large racetrack path with the CEAC prosthesis. Left: Overlay of pen-tip trajectories for Slow, Medium, and Fast trials; the inset (a) provides a magnified view. Right: Joint-angle time series for the Medium-speed trial. The horizontal ribbon marks intervals of active elbow (green) control ($\epsilon^{*}>0$). }
    \label{fig::Figure_5_CEAC_Drawing_illustrativeExample}
\end{figure}

In this subsection, we first show a single representative trial in which a participant performed the drawing task at the intended \emph{MEDIUM} speed on the large path (Fig.~\ref{fig::Figure_2_ExperimentalSetup}, panel C, 6) using the supernumerary CEAC-controlled prosthesis. 

The left panel of Fig. \ref{fig::Figure_5_CEAC_Drawing_illustrativeExample} shows the pen trajectory on the screen, demonstrating reduced precision at higher task speeds. The right panel of Figure \ref{fig::Figure_5_CEAC_Drawing_illustrativeExample} illustrates joint angles over time: elbow flexion $\beta$ (green), trunk flexion $\phi$ (red), and shoulder flexion $\theta$ (orange). To facilitate comparison between joints, all angles are expressed relative to their initial posture, where 0$^{\circ}$ corresponds to the starting angle. At trial onset, the trunk was initially flexed at approximately 76$^{\circ}$ (slightly forward-bent), the elbow at 106$^{\circ}$ (slightly extended), and the shoulder at 24$^{\circ}$.

Throughout this trial, the elbow actively contributes to the drawing task, mostly moving in coordination with the trunk and shoulder. The motor-driven elbow exhibits varying movement velocities, demonstrating continuous adaptation during the task. 

\subsubsection{Task performance during the drawing task}
\label{subsec:task_performance_drawing}

\begin{figure}[h]
    \centering
    \includegraphics[width=0.45\textwidth]{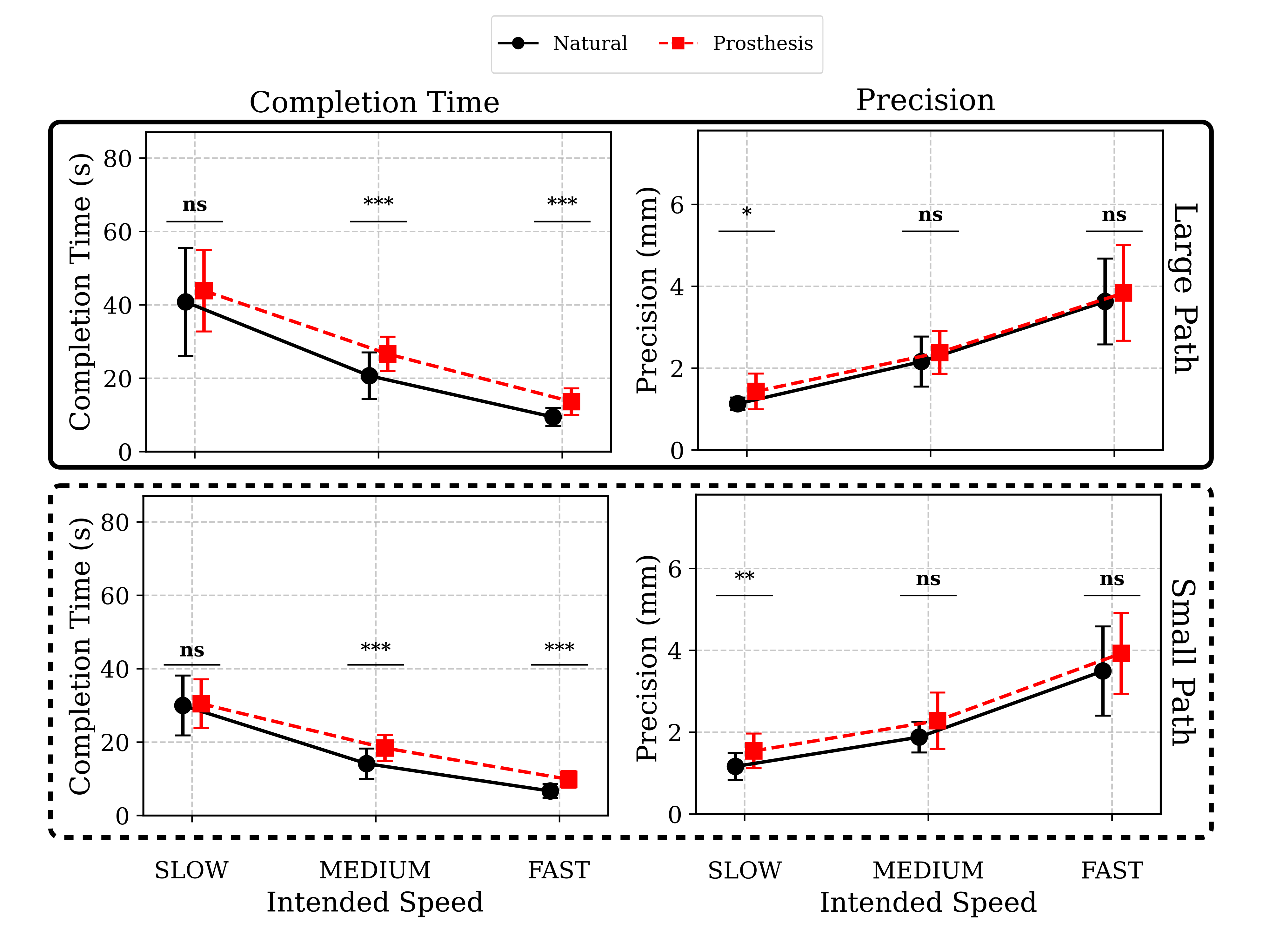}    
    \caption{Task-performance metrics for the drawing experiment. The figure is a $2 \times 2$ grid where columns show completion time and precision. The upper pair of panels (solid frame) correspond to the large path, while the lower pair (dashed frame) correspond to the small path. Data compares the natural-limb (black circles) and CEAC-prosthesis (red squares) conditions. Error bars represent one standard deviation.}
    \label{fig::Figure_6_Performance_parameters_drawing}
\end{figure}

Figure \ref{fig::Figure_6_Performance_parameters_drawing} summarizes completion time and geometric precision for the drawing experiment with 12 participants as a function of intended speed (\textit{SLOW}, \textit{MEDIUM}, \textit{FAST}) and path size (\textit{large}, \textit{small}).  
Across both path sizes, completion time shortened monotonically with the instructed speed for both the natural elbow and for the CEAC-controlled prosthesis.  
For the \textit{large} path, the prosthesis required more time than the natural elbow at \textit{MEDIUM} speed (natural: $20.68\pm6.36$\,s; prosthesis: $26.62\pm4.71$\,s; $p<0.001$) and at \textit{FAST} speed (natural: $9.45\pm2.47$\,s; prosthesis: $13.64\pm3.62$\,s; $p<0.001$); the difference was not significant at \textit{SLOW} speed ($p=0.42$).  
For the \textit{small} path the same pattern was observed: no difference at \textit{SLOW} speed ($p=0.78$), but longer completion times with the prosthesis at \textit{MEDIUM} ($14.14\pm4.11$\,s vs.\ $18.40\pm3.56$\,s; $p<0.001$) and \textit{FAST} speeds ($6.69\pm1.92$\,s vs.\ $9.86\pm2.29$\,s; $p<0.001$).

Precision decreased with movement speed for both natural and prosthesis, the difference, however, between natural and prosthesis in each condition was minimal.
For the \textit{large} path, precision differed only at \textit{SLOW} speed (natural: $1.13\pm0.15$\,mm; prosthesis: $1.43\pm0.44$\,mm; $p=0.041$); differences at \textit{MEDIUM} and \textit{FAST} speeds were not significant ($p=0.34$ and $p=0.36$, respectively).  
For the \textit{small} path, a significant difference appeared at \textit{SLOW} speed (natural: $1.17\pm0.33$\,mm; prosthesis: $1.54\pm0.42$\,mm; $p=0.0013$), whereas the groups did not differ at \textit{MEDIUM} ($p=0.092$) or \textit{FAST} speed ($p=0.085$).

\subsubsection{Kinematic strategy during the drawing task}
\label{subsec:kinematics_drawing}
\begin{figure}[h]
    \centering
    \includegraphics[width=0.45\textwidth]{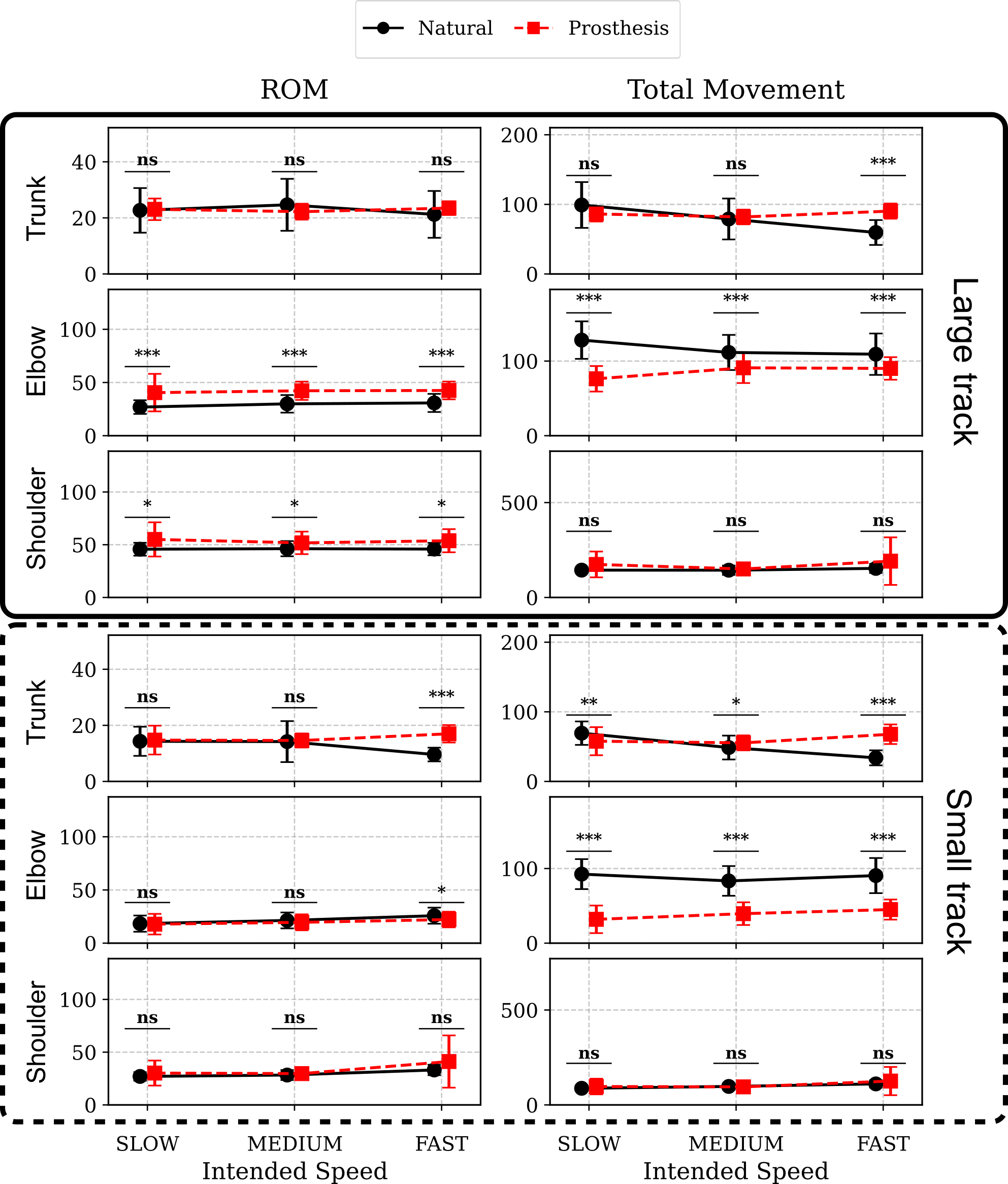}
    \caption{Joint kinematic metrics for the drawing task, comparing the natural-limb (black) and CEAC-prosthesis (red) conditions. The left column reports range of motion (ROM) and the right column reports total angular movement. The upper three rows (solid frame) correspond to the large track, while the lower three rows (dashed frame) correspond to the small track.}
    \label{fig::Figure_7_Posture_parameters_drawing}
\end{figure}

Figure~\ref{fig::Figure_7_Posture_parameters_drawing} indicates that elbow use changed markedly when participants operated the CEAC-controlled prosthesis. On the \textit{large} track the elbow range of motion was consistently greater with the prosthesis than with the natural elbow. For example, at the \textit{FAST} condition on the \textit{large} track the ROM was $30.8^{\circ}\!\pm\!8.5^{\circ}$ with the natural elbow and $42.5^{\circ}\!\pm\!8.4^{\circ}$ with the prosthesis, yet the cumulative elbow excursion over the whole path dropped from $109.2^{\circ}\!\pm\!27.7^{\circ}$ with the natural elbow to $90.2^{\circ}\!\pm\!15.1^{\circ}$ with the prosthesis. Thus, larger but fewer elbow sweeps were executed. Trunk range of motion on the large track remained comparable between conditions, whereas the total trunk movement increased only at \textit{FAST} speed in both the large and small tracks, from $59.6^{\circ}\!\pm\!17.9^{\circ}$ for the natural elbow to $90.3^{\circ}\!\pm\!10.9^{\circ}$ for the prosthesis.  Shoulder ROM and cumulative movement exhibited the same speed-dependent increase with the prosthesis and the natural elbow.

On the \textit{small} track, the pattern narrowed.  Elbow ROM differed little between prosthesis and natural elbow at \textit{SLOW} and \textit{MEDIUM} speeds and was slightly lower with the prosthesis at \textit{FAST} speed ($22.2^{\circ}\!\pm\!7.3^{\circ}$ versus $25.8^{\circ}\!\pm\!7.6^{\circ}$ with the natural elbow).  Nevertheless, the total elbow movement remained markedly smaller for the prosthesis over all speeds (e.g.\ \textit{FAST}: $45.0^{\circ}\!\pm\!13.5^{\circ}$ (prosthesis) versus $90.6^{\circ}\!\pm\!23.5^{\circ}$ (natural)).  The trunk total movement, in turn, nearly doubled at \textit{FAST} speed on the small track, from $33.9^{\circ}\!\pm\!10.8^{\circ}$ (natural) to $67.8^{\circ}\!\pm\!14.1^{\circ}$ (prosthesis), and trunk ROM climbed from $9.6^{\circ}\!\pm\!2.5^{\circ}$ (natural) to $17.0^{\circ}\!\pm\!3.1^{\circ}$ (prosthesis in the \textit{FAST} condition on the small track.  
Together these observations show that CEAC redistributes effort: the trunk provides additional excursion with respect to the natural gesture, allowing the prosthetic elbow to operate through wider but less frequent motions while shoulder activity rises modestly to accommodate the altered coordination.

\subsection{Reaching}

\subsubsection{Illustrative example of the Reaching Task}

\begin{figure}[!ht]
    \centering
    \includegraphics[width=0.35\textwidth]{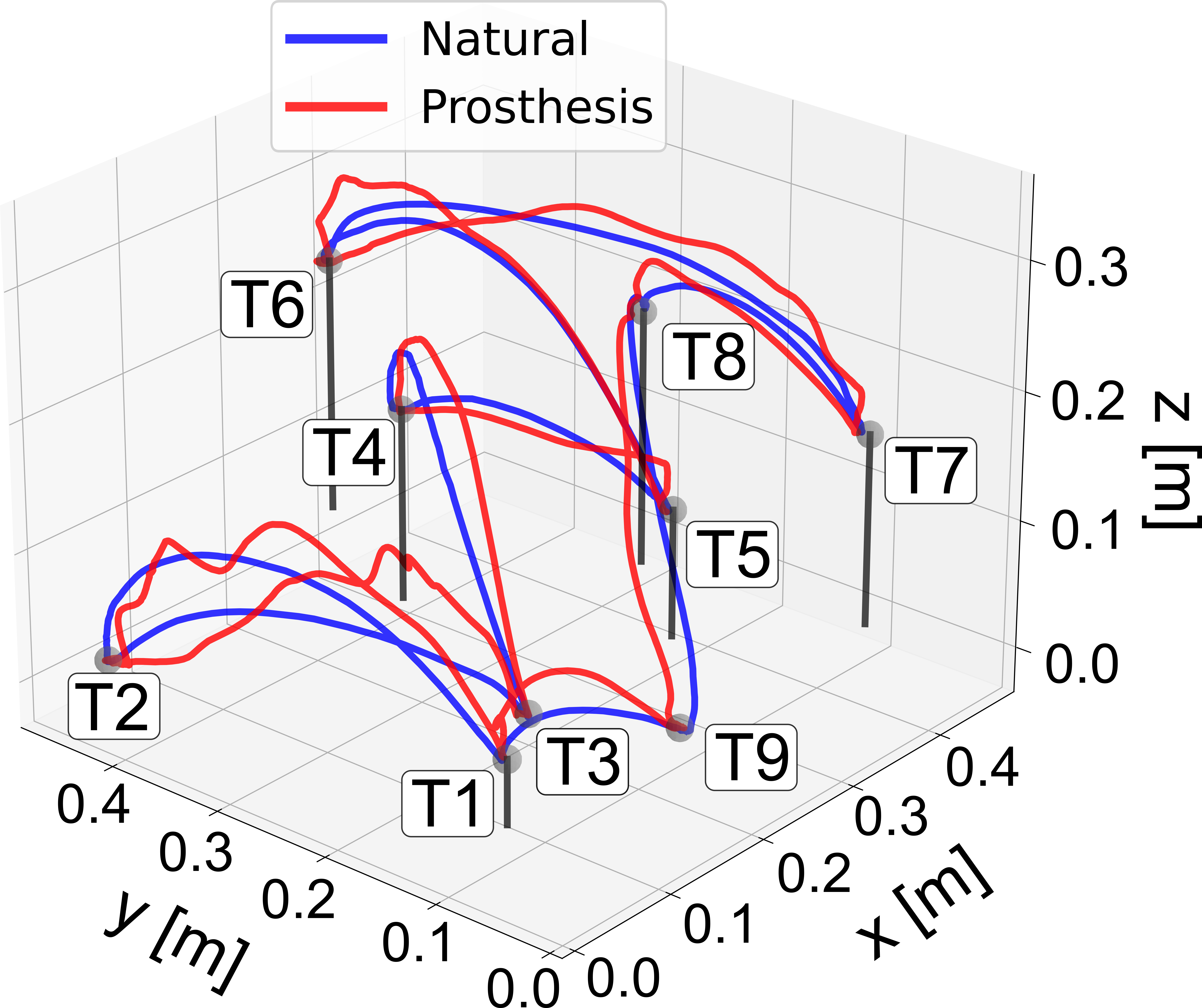}
    \caption{Pen-tip trajectories for the nine-target-reaching task. The path executed with the natural elbow is shown in blue; the path with the CEAC-controlled prosthesis is in red. Grey spheres and labels (T1–T9) denote the target sequence.}
    \label{fig::Figure_8_ReachingTask_3D}
\end{figure}
Figure~\ref{fig::Figure_8_ReachingTask_3D} shows the pen-tip trajectories recorded during one complete nine-target-reaching sequence. Blue traces correspond to the participant’s natural-limb movements, whereas red traces correspond to movements executed with the CEAC-controlled prosthesis. Grey spheres mark the centres of the tolerance volumes for targets T1–T9; thin vertical rods indicate each target’s distance from the screen. 

\subsubsection{Task performance in the reaching task}
\label{subsec:task_performance_reaching}
\begin{figure}[!ht]
    \centering
    \includegraphics[width=0.5\textwidth]{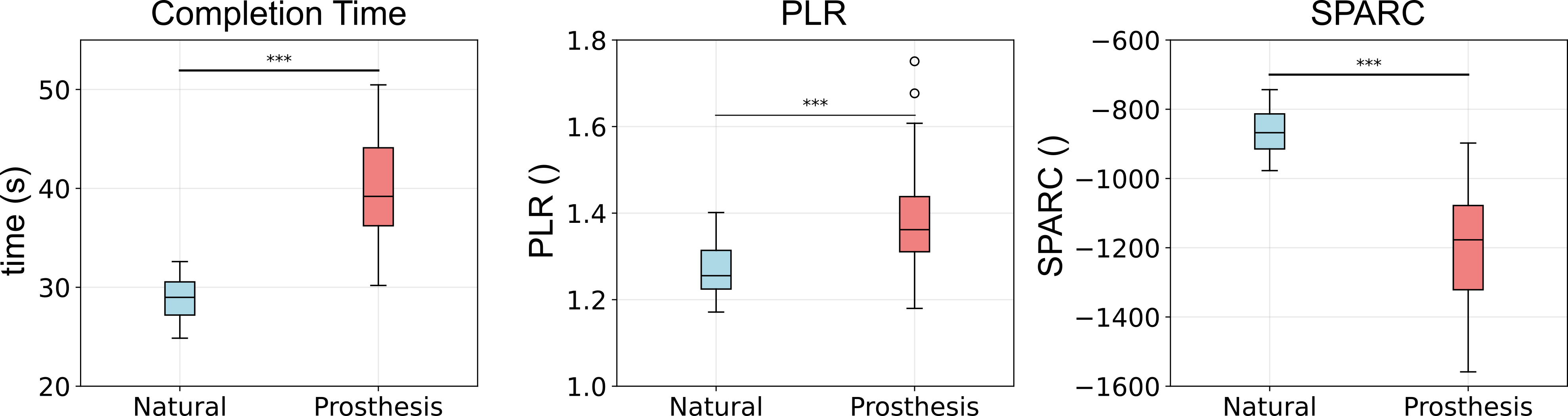}
    \caption{Performance metrics for the nine-target-reaching task, comparing the natural-limb (blue) and CEAC-prosthesis (red) conditions. Panels show, from left to right: task-completion time, path-length ratio (PLR), and spectral arc length (SPARC).}
    \label{fig::Figure_9_PerformanceParams_Reaching}
\end{figure}

Figure~\ref{fig::Figure_9_PerformanceParams_Reaching} compares outcome measures for the nine-target sequence executed by 10 participants with the CEAC-controlled prosthesis (red) and with the natural elbow (blue).
Completion time was longer with the prosthesis than with the natural elbow (\(39.9\!\pm\!5.4\)\,s versus \(28.9\!\pm\!2.2\)\,s; \(p<0.001\)).  
The path-length ratio likewise increased from \(1.27\!\pm\!0.10\) with the natural elbow to \(1.39\!\pm\!0.10\) (\(p<0.001\)) with the prosthesis, indicating less direct trajectories (as visible in Fig.~\ref{fig::Figure_8_ReachingTask_3D}).  
Spectral arc length of the tangential-speed profile became more negative with the prosthesis (\(-1200\!\pm\!175\) versus \(-865\!\pm\!66\); \(p<0.001\)) with the natural elbow, reflecting a higher frequency content and reduced smoothness.  

\subsubsection{Kinematic strategy in the reaching task}
\label{subsec:kinematics_reaching}

\begin{figure}[!ht]
    \centering
    \includegraphics[width=0.4\textwidth]{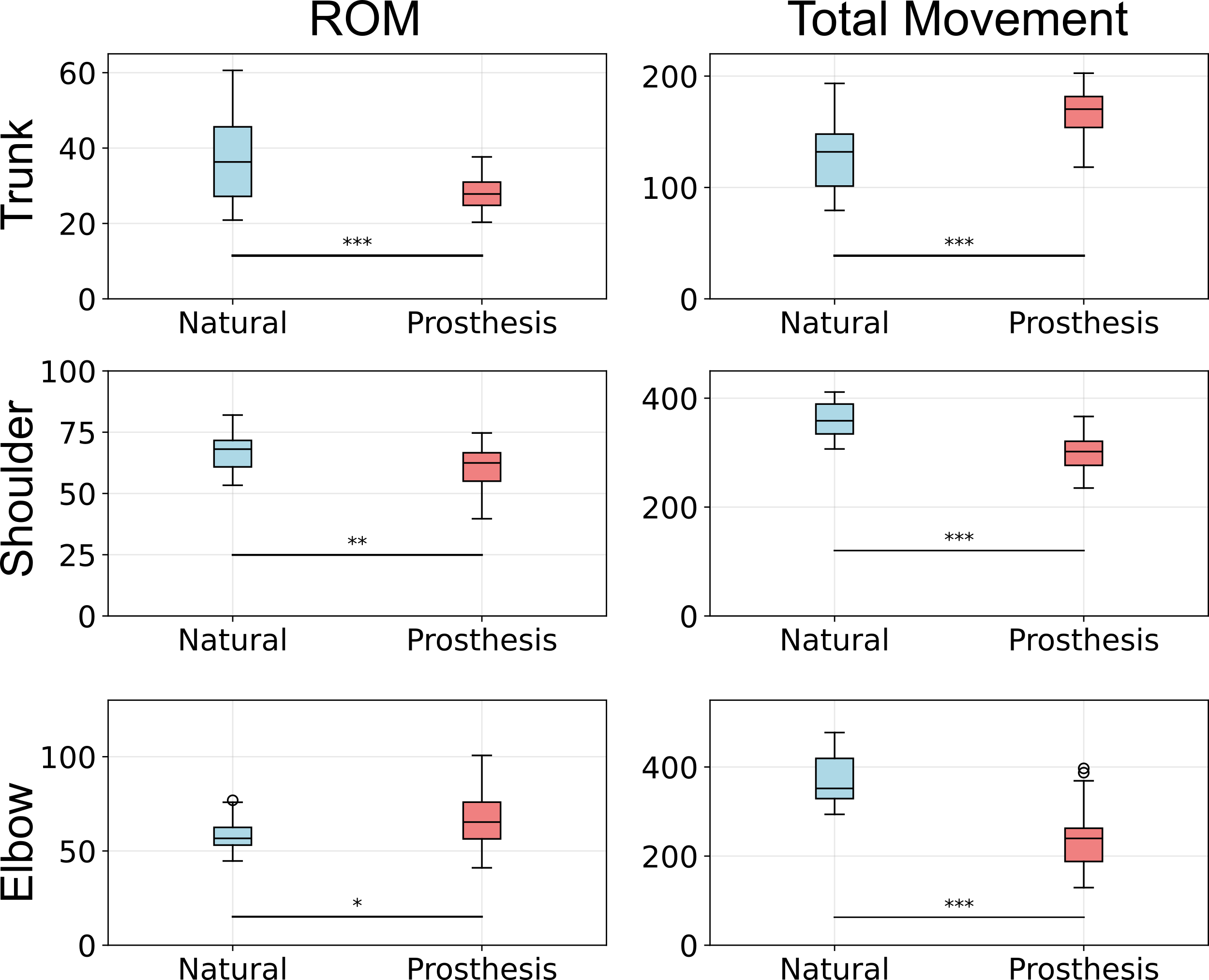}
    \caption{Body-segment kinematics during the reaching task. Columns show joint range of motion (ROM) and total angular movement. Rows compare the trunk, shoulder, and elbow kinematics for the natural-limb (blue) and prosthesis (red) conditions.}
    \label{fig::Figure_10_ReachingTask_JointKinematics}
\end{figure}

The joint-angle statistics in Figure~\ref{fig::Figure_10_ReachingTask_JointKinematics} show that control with the prosthesis altered the distribution of movement along the kinematic chain.  
Trunk range of motion decreased from \(37\!\pm\!10.7^{\circ}\) with the natural elbow to \(28\!\pm\!4.5^{\circ}\) with the prosthesis (\(p<0.001\)), yet the cumulative trunk excursion rose (\(129\!\pm\!29.4^{\circ}\) versus \(167\!\pm\!20.5^{\circ}\); \(p<0.001\)), indicating more frequent trunk adjustments of smaller magnitude.  
At the elbow joint level, the opposite trend emerged: ROM increased (\(66\!\pm\!14.1^{\circ}\) with the prosthesis versus \(59\!\pm\!8.2^{\circ}\); \(p=0.03\)) while the total angular change fell (\(238\!\pm\!64.8^{\circ}\) versus \(371\!\pm\!55.6^{\circ}\); (p<0.001)), showing that fewer, larger elbow sweeps of the prosthesis replaced the many small corrections observed with the natural elbow.  
Shoulder curves (not tabulated) followed the elbow pattern, exhibiting a moderate rise in ROM and a concomitant fall in cumulative excursion when the prosthesis was used.

To quantify the coordination between joints, we computed the Synergy Joint Velocity Index (SJVI) from shoulder and elbow velocities during the inter-target movements. This index was significantly lower when using the prosthesis compared to the natural elbow \(\left(0.60 \pm 0.05 \,\text{vs.}\, 0.80 \pm 0.04,\ p < 0.001\right)\)\, confirming a slightly looser temporal coupling between shoulder and elbow joint velocities under CEAC.
\vspace{2cm}

\section{Discussion}
\label{sec:discussion}

In this study, we introduced and preliminarily validated Compensation Effect Amplification Control (CEAC), a novel movement-based control paradigm for active elbow prostheses. Our results, gathered from complex path tracking and reaching tasks, demonstrate that CEAC enables effective, coordinated control of the prosthesis, achieving task performance that is largely comparable to that of a natural elbow, without requiring extreme compensatory movements or significant changes in motor strategy.

Our preliminary investigation of a simple line-drawing task (Fig.~\ref{fig::Figure_3_CCCvsCEACvsNatural_exp}) confirmed our central hypothesis regarding the limitations of existing approaches and the benefits of CEAC. As anticipated (Fig.~\ref{fig::Figure_1_CCC_vs_CEAC_sketch}), Compensation Cancellation Control (CCC), with its fixed upright reference, proved ill-suited for the continuous tracking task, forcing the user into a two-phase, non-intuitive manoeuvre involving target overshooting and ergonomically challenging backward bends to halt or reverse the motion. In stark contrast, CEAC's dynamic reference posture allowed the user's trunk to act as a functional input. This allows the trunk from being a purely compensatory segment once again as an active contributor to the task, effectively increasing the user's functional workspace and enabling a smooth, continuous trajectory and additionally, closely resembling the joint kinematics of the natural limb.

A key strength of CEAC demonstrated in this study is its adaptability across different task using a single, fixed set of control parameters. The speed modulation experiment (Fig.~\ref{fig::Figure_4_CEAC_5speeds_exp}) showed that participants could intuitively modulate their end-effector velocity from \textasciitilde 3~cm/s to \textasciitilde 30~cm/s simply by adjusting the amplitude and timing of their trunk movements, leveraging the "catch-release" dynamic inherent to the controller. While control was lost at the extremely slow speed of 1.5~cm/s, this was an expected outcome of a condition where trunk velocity remained below the minimal speed $|\dot{\phi}_{0,\max}|$, making the task entirely compensatory. The ability to successfully perform a complex drawing task (Fig.~\ref{fig::Figure_5_CEAC_Drawing_illustrativeExample}), characterized by continuous changes in direction and speed, further underscores the controller's capacity to manage time-varying velocity commands and facilitate simultaneous, coordinated motion of the trunk, shoulder, and prosthetic elbow.

When compared against the benchmark of a asymptomatic natural limb, CEAC-based control demonstrated promising efficacy. In the drawing task (Fig.~\ref{fig::Figure_6_Performance_parameters_drawing}), although completion times were modestly longer for the prosthesis at higher speeds, the absolute difference in precision was minimal and often not statistically significant. The underlying kinematic strategies (Fig.~\ref{fig::Figure_7_Posture_parameters_drawing}) reveal a difference of CEAC to the natural limb: a redistribution of motor effort. For instance, on the large path, participants exhibited a larger elbow ROM with the prosthesis (e.g., 42.5$^{\circ}$~$\pm$~8.4$^{\circ}$ vs. 30.8$^{\circ}$~$\pm$~8.5$^{\circ}$ for natural elbow at FAST speed) yet a \textit{smaller} total cumulative elbow movement (90.2$^{\circ}$~$\pm$~15.1$^{\circ}$ vs. 109.2$^{\circ}$~$\pm$~27.7$^{\circ}$). This indicates that CEAC encourages fewer and larger elbow sweeps, as opposed to the natural limb.

This trend was also evident in the 3D reaching task. While performance was, as expected, slower and less smooth than with the natural elbow (Fig. \ref{fig::Figure_9_PerformanceParams_Reaching}), the control was effective and the kinematic trade-offs were consistent (Fig. \ref{fig::Figure_10_ReachingTask_JointKinematics}). The trunk was used more frequently (higher total movement) but not to a greater extent (similar ROM), indicating more active participation without over-compensation. Crucially, the coordinated action between joints remained high. The Synergy Joint Velocity Index (SJVI) of 0.60, while lower than the natural limb's 0.80, still demonstrates a substantial level of simultaneous joint activation. This is a stark contrast to the sequential, "joint-by-joint" movements often observed with conventional myoelectric prosthesis and highlights the potential of CEAC to restore more holistic movement patterns.

The use of a VIVE Ultimate Tracker provided robust and reliable input for this study, creating a system that functions independently of laboratory motion capture. For the current single-joint control scheme, a simpler Inertial Measurement Unit (IMU) would have sufficed to measure trunk orientation. However, our choice was deliberate, as the absolute position and orientation data available from the VIVE tracker will be valuable for future work extending CEAC to simultaneously control multiple prosthetic degrees of freedom, where the relative positions of different body segments become critical. 

Future work will also aim to develop adaptive gain mechanisms to better tailor system responsiveness to specific tasks and individual user preferences. In parallel, the performance of CEAC will have to be benchmarked against conventional EMG-based control strategies in participants with UL amputations. Particular attention should be paid to evaluating performance in UL amputees wearing prostheses with sockets and straps, as these components may constrain certain movement capabilities. Moreover, future research should address the current limitations of the CEAC system by investigating potential integration with myoelectric control (particularly for the control of prosthetic hand grasping), extending the approach to multi-joint control, and developing strategies to detect and leverage body compensations during complex, ecologically valid tasks—such as ambulation. Finally evaluating user experience and gathering their feedback with a questionnaire such as NASA TLX could allow us to deepen the understanding of the cognitive load and the intuitiveness of the CEAC.  

\section{Conclusion}
\label{sec:conclusion}

This paper introduced Compensation Effect Amplification Control (CEAC), a versatile movement-based paradigm for the continuous position and velocity control of an intermediate prosthetic joint. Building on the framework of compensation-based control, CEAC incorporates a dynamic reference that follows the user's trunk orientation with a controlled delay. This unique approach allows the trunk to be used as a functional input, enabling intuitive, continuous, and velocity-modulated control of the prosthetic elbow.

We preliminarily validated this strategy in drawing and reaching experiments, comparing the performance of able-bodied participants using a supernumerary prosthesis against the benchmark of their own natural limb. The results demonstrated that users could effectively coordinate the prosthesis with their natural joints to perform complex tasks requiring precise and continuous trajectory control. CEAC enabled task performance and kinematic strategies that were comparable to natural movement, fostering a high degree of simultaneous joint motion. By purposefully leveraging rather than simply canceling natural body movements, CEAC represents a promising and intuitive approach for overcoming the limitations of existing control methods and improving the quality of life of UL prosthesis users. 

\bibliographystyle{IEEEtran}
\bibliography{references}

\begin{thebibliography}{10}
\providecommand{\url}[1]{#1}
\csname url@samestyle\endcsname
\providecommand{\newblock}{\relax}
\providecommand{\bibinfo}[2]{#2}
\providecommand{\BIBentrySTDinterwordspacing}{\spaceskip=0pt\relax}
\providecommand{\BIBentryALTinterwordstretchfactor}{4}
\providecommand{\BIBentryALTinterwordspacing}{\spaceskip=\fontdimen2\font plus
\BIBentryALTinterwordstretchfactor\fontdimen3\font minus
  \fontdimen4\font\relax}
\providecommand{\BIBforeignlanguage}[2]{{%
\expandafter\ifx\csname l@#1\endcsname\relax
\typeout{** WARNING: IEEEtran.bst: No hyphenation pattern has been}%
\typeout{** loaded for the language `#1'. Using the pattern for}%
\typeout{** the default language instead.}%
\else
\language=\csname l@#1\endcsname
\fi
#2}}
\providecommand{\BIBdecl}{\relax}
\BIBdecl

\bibitem{abayasiri2017mobio}
R.~A.~M. Abayasiri, D.~K. Madusanka, N.~Arachchige, A.~Silva, and R.~Gopura,
  ``Mobio: A 5 dof trans-humeral robotic prosthesis,'' in \emph{2017
  International Conference on Rehabilitation Robotics (ICORR)}.\hskip 1em plus
  0.5em minus 0.4em\relax IEEE, 2017, pp. 1627--1632.

\bibitem{resnik2014deka}
L.~Resnik, S.~L. Klinger, and K.~Etter, ``The deka arm: Its features,
  functionality, and evolution during the veterans affairs study to optimize
  the deka arm,'' \emph{Prosthetics and orthotics international}, vol.~38,
  no.~6, pp. 492--504, 2014.

\bibitem{alshammary2017synergistic}
N.~A. Alshammary, D.~A. Bennett, and M.~Goldfarb, ``Synergistic elbow control
  for a myoelectric transhumeral prosthesis,'' \emph{IEEE Transactions on
  Neural Systems and Rehabilitation Engineering}, vol.~26, no.~2, pp. 468--476,
  2017.

\bibitem{farina2014extraction}
D.~Farina, N.~Jiang, H.~Rehbaum, A.~Holobar, B.~Graimann, H.~Dietl, and O.~C.
  Aszmann, ``The extraction of neural information from the surface emg for the
  control of upper-limb prostheses: emerging avenues and challenges,''
  \emph{IEEE Transactions on Neural Systems and Rehabilitation Engineering},
  vol.~22, no.~4, pp. 797--809, 2014.

\bibitem{fougner2012control}
A.~Fougner, {\O}.~Stavdahl, P.~J. Kyberd, Y.~G. Losier, and P.~A. Parker,
  ``Control of upper limb prostheses: Terminology and proportional myoelectric
  control—a review,'' \emph{IEEE Transactions on neural systems and
  rehabilitation engineering}, vol.~20, no.~5, pp. 663--677, 2012.

\bibitem{brack2021review}
R.~Brack and E.~H. Amalu, ``A review of technology, materials and r\&d
  challenges of upper limb prosthesis for improved user suitability,''
  \emph{Journal of Orthopaedics}, vol.~23, pp. 88--96, 2021.

\bibitem{trent2020narrative}
L.~Trent, M.~Intintoli, P.~Prigge, C.~Bollinger, L.~S. Walters, D.~Conyers,
  J.~Miguelez, and T.~Ryan, ``A narrative review: current upper limb prosthetic
  options and design,'' \emph{Disability and Rehabilitation: Assistive
  Technology}, 2020.

\bibitem{connan2016assessmentEMG}
M.~Connan, E.~Ruiz~Ram{\'\i}rez, B.~Vodermayer, and C.~Castellini, ``Assessment
  of a wearable force-and electromyography device and comparison of the related
  signals for myocontrol,'' \emph{Frontiers in neurorobotics}, vol.~10, p.~17,
  2016.

\bibitem{jiang2012emg}
N.~Jiang, J.~L. Vest-Nielsen, S.~Muceli, and D.~Farina, ``Emg-based
  simultaneous and proportional estimation of wrist/hand kinematics in
  uni-lateral trans-radial amputees,'' \emph{Journal of neuroengineering and
  rehabilitation}, vol.~9, pp. 1--11, 2012.

\bibitem{hahne2014linear}
J.~M. Hahne, F.~Biessmann, N.~Jiang, H.~Rehbaum, D.~Farina, F.~C. Meinecke,
  K.-R. M{\"u}ller, and L.~C. Parra, ``Linear and nonlinear regression
  techniques for simultaneous and proportional myoelectric control,''
  \emph{IEEE Transactions on Neural Systems and Rehabilitation Engineering},
  vol.~22, no.~2, pp. 269--279, 2014.

\bibitem{smith2015evaluation}
L.~H. Smith, T.~A. Kuiken, and L.~J. Hargrove, ``Evaluation of linear
  regression simultaneous myoelectric control using intramuscular emg,''
  \emph{IEEE Transactions on Biomedical Engineering}, vol.~63, no.~4, pp.
  737--746, 2015.

\bibitem{jiang2013accurate}
N.~Jiang, I.~Vujaklija, H.~Rehbaum, B.~Graimann, and D.~Farina, ``Is accurate
  mapping of emg signals on kinematics needed for precise online myoelectric
  control?'' \emph{IEEE Transactions on Neural Systems and Rehabilitation
  Engineering}, vol.~22, no.~3, pp. 549--558, 2013.

\bibitem{farina2024EMGImpedance}
L.~Ferrante, M.~Sridharan, C.~Zito, and D.~Farina, ``Toward impedance control
  in human-machine interfaces for upper-limb prostheses,'' \emph{IEEE
  Transactions on Biomedical Engineering}, 2024.

\bibitem{latash2007MotorSynergies}
M.~L. Latash, J.~P. Scholz, and G.~Sch{\"o}ner, ``Toward a new theory of motor
  synergies,'' \emph{Motor control}, vol.~11, no.~3, pp. 276--308, 2007.

\bibitem{santello2013neural}
M.~Santello, G.~Baud-Bovy, and H.~J{\"o}rntell, ``Neural bases of hand
  synergies,'' \emph{Frontiers in computational neuroscience}, vol.~7, p.~23,
  2013.

\bibitem{Merad2020}
M.~Merad, N.~Jarrasse, E.~D. Montalivet, M.~Legrand, E.~Mastinu,
  M.~Ortiz-Catalan, A.~Touillet, N.~Martinet, J.~Paysant, and A.~Roby-Brami,
  ``Assessment of an automatic prosthetic elbow control strategy using residual
  limb motion for transhumeral amputated individuals with socket or
  osseointegrated prostheses,'' \emph{IEEE Transactions on Medical Robotics and
  Bionics}, vol.~2, pp. 38--49, 2 2020.

\bibitem{Mick2021}
S.~Mick, E.~Segas, L.~Dure, C.~Halgand, J.~Benois-Pineau, G.~E. Loeb,
  D.~Cattaert, and A.~de~Rugy, ``Shoulder kinematics plus contextual target
  information enable control of multiple distal joints of a simulated
  prosthetic arm and hand,'' \emph{Journal of NeuroEngineering and
  Rehabilitation}, vol.~18, 12 2021.

\bibitem{Segas2023}
E.~Segas, S.~Mick, V.~Leconte, O.~Dubois, R.~Klotz, D.~Cattaert, and
  A.~de~Rugy, ``Intuitive movement-based prosthesis control enables arm
  amputees to reach naturally in virtual reality,'' \emph{eLife}, vol.~12, 10
  2023.

\bibitem{khoramshahi2021intent}
M.~Khoramshahi, G.~Morel, and N.~Jarrasse, ``Intent-aware control in
  kinematically redundant systems: Towards collaborative wearable robots,'' in
  \emph{2021 IEEE International Conference on Robotics and Automation
  (ICRA)}.\hskip 1em plus 0.5em minus 0.4em\relax IEEE, 2021, pp.
  10\,453--10\,460.

\bibitem{GarciaRosas2020}
R.~Garcia-Rosas, D.~Oetomo, C.~Manzie, Y.~Tan, and P.~Choong, ``Task-space
  synergies for reaching using upper-limb prostheses,'' \emph{IEEE Transactions
  on Neural Systems and Rehabilitation Engineering}, vol.~28, pp. 2966--2977,
  12 2020.

\bibitem{Haddadin2024}
\BIBentryALTinterwordspacing
J.~Kühn, T.~Hu, A.~Tödtheide, E.~P. Fortunić, E.~Jensen, and S.~Haddadin,
  ``The synergy complement control approach for seamless limb-driven
  prostheses,'' \emph{Nature Machine Intelligence}, 4 2024. [Online].
  Available: \url{https://www.nature.com/articles/s42256-024-00825-7}
\BIBentrySTDinterwordspacing

\bibitem{Legrand2021}
M.~Legrand, N.~Jarrasse, E.~D. Montalivet, F.~Richer, and G.~Morel, ``Closing
  the loop between body compensations and upper limb prosthetic movements: A
  feasibility study,'' \emph{IEEE Transactions on Medical Robotics and
  Bionics}, vol.~3, pp. 230--240, 2 2021.

\bibitem{feder2024general}
M.~Feder, G.~Grioli, M.~G. Catalano, and A.~Bicchi, ``A general compensation
  control method for human--robot integration,'' \emph{The International
  Journal of Robotics Research}, p. 02783649251344636, 2024.

\bibitem{legrand2020wireloop}
M.~Legrand, N.~Jarrass{\'e}, F.~Richer, and G.~Morel, ``A closed-loop and
  ergonomic control for prosthetic wrist rotation,'' in \emph{2020 IEEE
  International Conference on Robotics and Automation (ICRA)}.\hskip 1em plus
  0.5em minus 0.4em\relax IEEE, 2020, pp. 2763--2769.

\bibitem{metzger2012characterization}
A.~J. Metzger, A.~W. Dromerick, R.~J. Holley, and P.~S. Lum, ``Characterization
  of compensatory trunk movements during prosthetic upper limb reaching
  tasks,'' \emph{Archives of physical medicine and rehabilitation}, vol.~93,
  no.~11, pp. 2029--2034, 2012.

\bibitem{MyRepo}
{Anonymous}, ``{Ultimate Tracker Python},''
  \url{https://anonymous.4open.science/r/UltimateTracker\_python-BB0D}, 2024,
  gitHub repository.

\bibitem{balasubramanian2015analysis}
S.~Balasubramanian, A.~Melendez-Calderon, A.~Roby-Brami, and E.~Burdet, ``On
  the analysis of movement smoothness,'' \emph{Journal of neuroengineering and
  rehabilitation}, vol.~12, pp. 1--11, 2015.

\end{thebibliography}

\end{document}